\documentclass[10pt,twocolumn,letterpaper,table,xcdraw]{article}

\usepackage{cvpr}              

\usepackage{booktabs}
\usepackage{multirow}
\usepackage{pifont}
\usepackage{tcolorbox}
\usepackage{listings}

\definecolor{cvprblue}{rgb}{0.21,0.49,0.74}
\usepackage[pagebackref,breaklinks,colorlinks,allcolors=cvprblue]{hyperref}

\def\paperID{4796} 
\def\confName{CVPR}
\def\confYear{2026}

\title{REVISOR: Beyond Textual Reflection, Towards Multimodal Introspective Reasoning in Long-Form Video Understanding}

\author{Jiaze Li\textsuperscript{1\ddag}\thanks{Project leader. \textsuperscript{\ddag} These authors contributed equally.}, Hao Yin\textsuperscript{1\ddag}, Wenhui Tan\textsuperscript{3\ddag}, Jingyang Chen\textsuperscript{1\ddag}, Boshen Xu\textsuperscript{3} \\ Yuxun Qu\textsuperscript{1}, Yijing Chen\textsuperscript{3}, Jianzhong Ju\textsuperscript{1}\thanks{Corresponding author.}, Zhenbo Luo\textsuperscript{1}, Jian Luan\textsuperscript{1} \\
  \textsuperscript{1}MiLM Plus, Xiaomi Inc.  
  \textsuperscript{3}Renmin University of China \\
}

\begin{document}
\maketitle
\begin{abstract}
Self-reflection mechanisms that rely on purely text-based rethinking processes perform well in most multimodal tasks. However, when directly applied to long-form video understanding scenarios, they exhibit clear limitations. The fundamental reasons for this lie in two points: (1) long-form video understanding involves richer and more dynamic visual input, meaning rethinking only the text information is insufficient and necessitates a further rethinking process specifically targeting visual information; (2) purely text-based reflection mechanisms lack cross-modal interaction capabilities, preventing them from fully integrating visual information during reflection. Motivated by these insights, we propose REVISOR (REflective VIsual Segment Oriented Reasoning), a novel framework for tool-augmented multimodal reflection. REVISOR enables MLLMs to collaboratively construct introspective reflection processes across textual and visual modalities, significantly enhancing their reasoning capability for long-form video understanding. To ensure that REVISOR can learn to accurately review video segments highly relevant to the question during reinforcement learning, we designed the Dual Attribution Decoupled Reward (DADR) mechanism. Integrated into the GRPO training strategy, this mechanism enforces causal alignment between the model’s reasoning and the selected video evidence. Notably, the REVISOR framework significantly enhances long-form video understanding capability of MLLMs without requiring supplementary supervised fine-tuning or external models, achieving impressive results on four benchmarks including VideoMME, LongVideoBench, MLVU, and LVBench. 

\end{abstract}    
\section{Introduction}
\label{sec:introduction}

Multimodal reasoning is fundamental to many real-world applications, such as interpreting scientific figures, performing geometric reasoning, and solving complex vision–language understanding tasks. However, when reinforcement learning (RL)-based reasoning strategies \citep{guo2025deepseek} are transferred from text-only models to multimodal settings \citep{meng2025mm,shen2025vlm,yang2025r1}, they often fail to show clear advantages over “fast-thinking” models and may even underperform in certain cases. This phenomenon primarily arises because current multimodal large language models (MLLMs) typically generate outputs under a token-level Markov assumption \citep{yu2025dapo,long2023large}, relying on local contextual dependencies. Such locality often leads to repetitive or incorrect reasoning steps \citep{zhang2025entropy}. Recent studies suggest that incorporating a self-reflection mechanism \citep{madaan2023self,kumar2024training} can mitigate these issues. By explicitly guiding the model to review, evaluate, and revise its reasoning trajectory, self-reflection helps prune invalid or erroneous reasoning paths, improve logical consistency, and promote deeper multimodal understanding\citep{gandhi2025cognitive,wan2025srpo}.

Despite significant progress in reflection mechanisms, most studies rely on text-only reconsideration processes. Such methods perform well on general multimodal tasks. However, when directly applied to long-form video understanding, their limitations become evident. As demonstrated in our experiments in \cref{subsec:dilemma}, the representative text-based reflection approach, VL-Rethinker \citep{wang2025vl}, leads to degraded performance in long-form video scenarios. We attribute this phenomenon to two main factors: 


\begin{tcolorbox}[colback=orange!10, colframe=orange!65!black, boxrule=0.5mm]
\begin{itemize}
    \item \textit{Unlike image understanding, long-form video reasoning involves richer and more dynamic visual inputs. Purely text-based reflection is insufficient to correct reasoning errors without explicitly reconsidering visual information.}
    \vspace{0.1cm}
    \item \textit{Text-only reflection lack cross-modal interaction capabilities, preventing the model from integrating visual cues during reflection. This limitation restricts the reasoning improvement potential of MLLMs in video understanding tasks.}
\end{itemize}
\end{tcolorbox}

\begin{figure*}[t]
  \centering
   \includegraphics[width=0.95\linewidth]{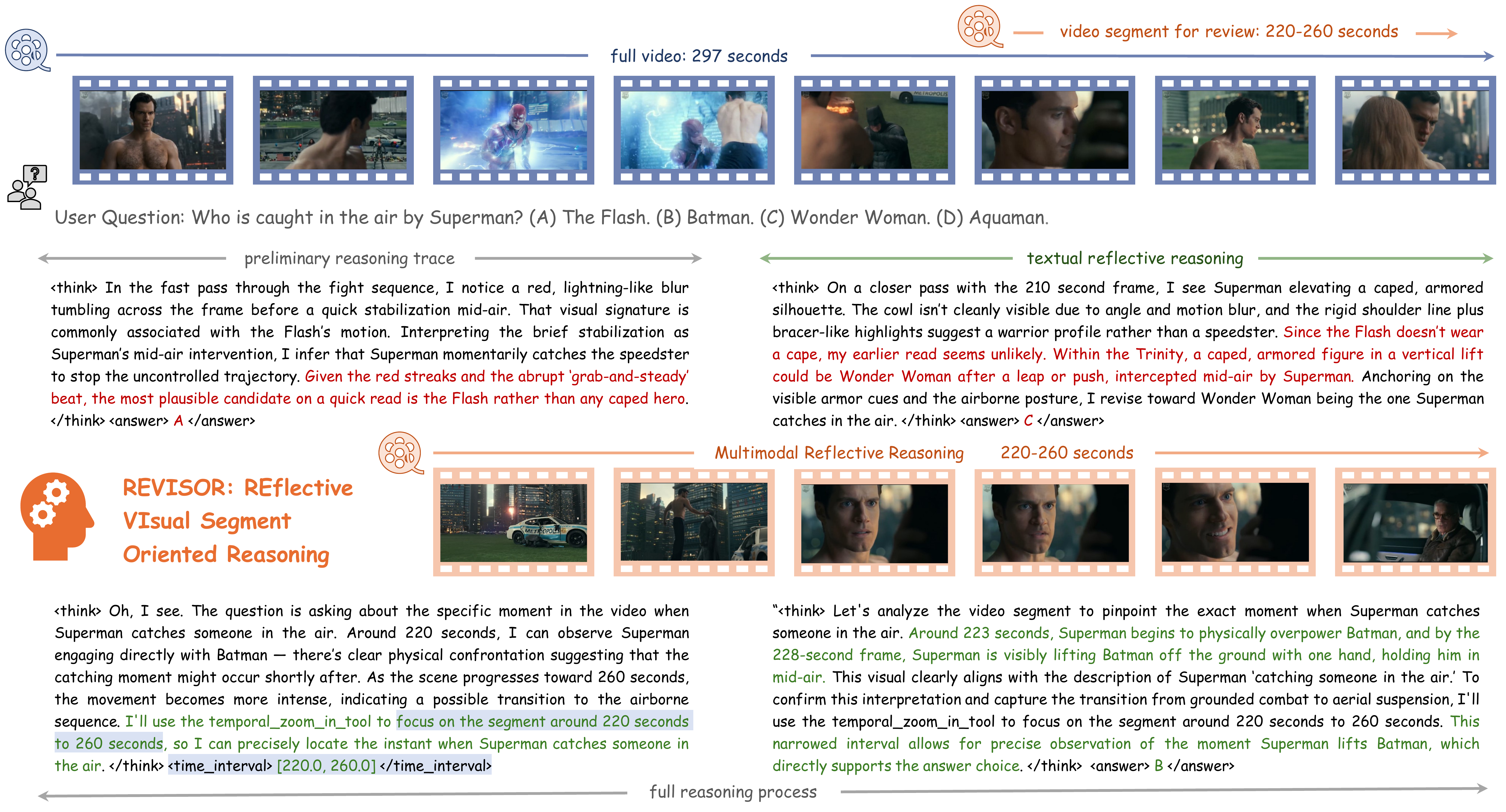}
   \caption{Operational workflow of the proposed REVISOR framework, contrasting it with traditional reflection mechanisms. The top panel illustrates a typical traditional approach, often employing a text-based re-evaluation mechanism. In contrast, the bottom panel details the REVISOR framework. This process involves two distinct stages: (1) Initial Inference, which generates a preliminary reasoning trace and identifies critical regions for detailed analysis; and (2) Reflective Reasoning, which integrates this initial trace with newly sampled, fine-grained visual evidence to yield a refined and robust final prediction.}
   \label{fig: method revisor}
\end{figure*}


Motivated by the insights above, we propose REVISOR (REflective VIsual Segment Oriented Reasoning), a novel two-stage reasoning framework designed to enhance video understanding. REVISOR leverages a tool-augmented multimodal reflection mechanism that enables MLLMs to collaboratively construct introspective reflection processes across textual and visual modalities, thereby enhancing reasoning capability. As illustrated in \cref{fig: method revisor}, REVISOR comprises an MLLM that collaborates with a visual toolbox. In Stage 1, the MLLM performs an initial reasoning step, identifies video segments requiring further examination, and invokes the visual toolbox to resample key frames from these segments as supplementary inputs. In Stage 2, the MLLM integrates the initial reasoning trace with the reviewed visual frames to iteratively refine its reasoning, ultimately generating a more accurate response.

\begin{figure*}[t]
  \centering
   \includegraphics[width=0.95\linewidth]{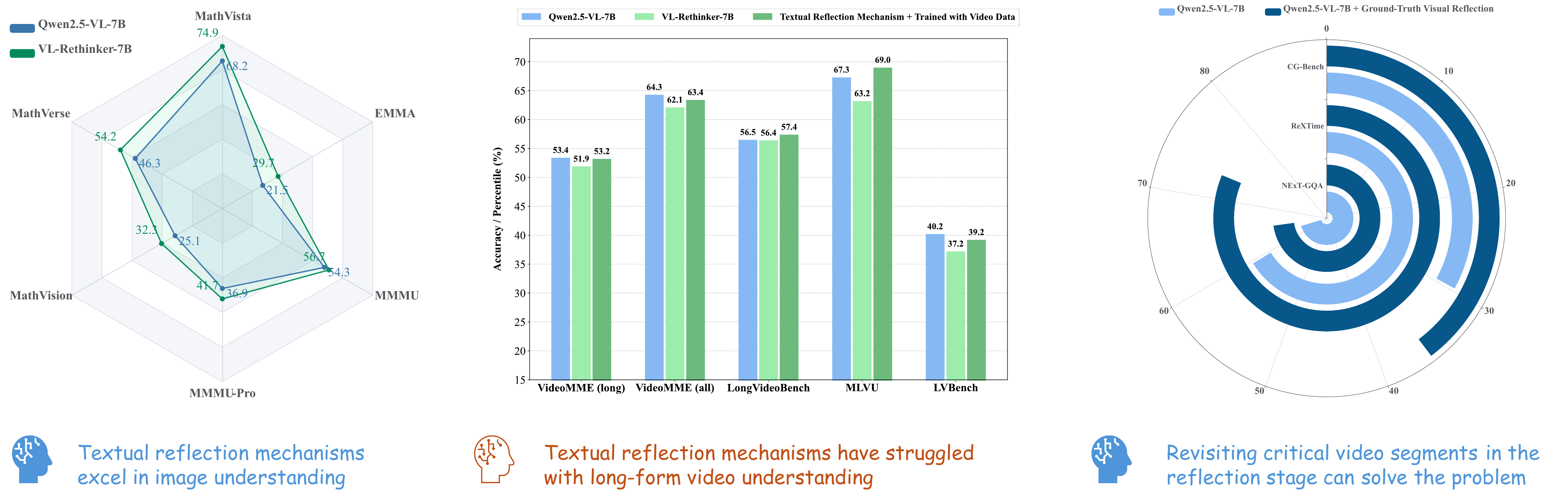}
   \caption{Motivation for proposing a multimodal reflection mechanism. Left: Text-only reflection mechanisms, such as VL-Rethinker, achieve significant performance improvements in image understanding tasks. Middle: However, applying the same text-based reflection strategy to long-form video understanding leads to performance degradation. Right: Incorporating a revisit of key video segments during the reflection stage effectively improves performance on video understanding tasks.}
   \label{fig: motivation}
\end{figure*}

Relying solely on verification-based reinforcement from the final answer during GRPO \citep{shao2024deepseekmath} training constrains REVISOR’s capacity to identify question-relevant review segments in its reflection stage. To overcome this limitation, we propose a Dual Attribution Decoupled Reward (DADR) mechanism in \cref{subsec:reward}, which supplements the final-answer verification reward with a Causal Segment Sufficiency Reward (CSSR). The CSSR enforces causal alignment between the model’s reasoning and the selected video evidence, rewarding correctness only when the answer is derived exclusively from those segments. This mechanism explicitly encourages the model to focus on informative visual cues while implicitly discouraging reliance on irrelevant or spurious content, thereby enhancing REVISOR’s multimodal reasoning consistency and overall robustness.

The REVISOR framework significantly enhances the video understanding capabilities of MLLMs without requiring supplementary supervised fine-tuning (SFT) or external models. Across the VideoMME \citep{fu2025video}, LongVideoBench \citep{wu2024longvideobench}, MLVU \citep{zhou2025mlvu}, and LVBench \citep{wang2025lvbench} benchmarks, REVISOR consistently improves the base model's average accuracy by about 2\%. Extended analysis in \cref{sec:discussion} validates two key drivers of REVISOR's success: (1) \textit{for long-form video understanding, revisiting and refining visual information is more critical than rethinking the textual reasoning process}; and (2) \textit{DADR mechanism ensures the precise recall and utilization of salient visual cues}.

In summary, the contributions of this paper are threefold:
\begin{itemize}
    \item We diagnose the root cause for the poor performance of conventional text-based self-reflection mechanisms when applied to long-form video understanding.
    \item We propose REVISOR, a two-stage reasoning framework that transforms the conventional text-based self-reflection mechanism into a tool-augmented multimodal one, significantly enhancing the model's reasoning capability and accuracy in long-form video understanding tasks.
    \item We integrate the DADR mechanism into the GRPO method, enabling MLLMs to learn how to review the correct video content during self-reflection within the RL training process, thereby maximizing REVISOR's performance boost on long-form video comprehension.
\end{itemize}
\section{Motivation}
\label{sec:motivation}

\cref{subsec:dilemma} shows that text-only reflection mechanisms are inadequate for long-form video understanding. The underlying reason is that, unlike static images, long-form videos present far richer and more dynamic visual information, making purely textual re-evaluation insufficient for correcting reasoning errors. In \cref{subsec:necessity}, we initially demonstrated the importance of incorporating a visual re-evaluation process during the reflection stage, with preliminary experiments showing that this significantly enhances the accuracy of MLLMs in long-form video understanding tasks.

\subsection{Dilemma of Text-Based Reflection}
\label{subsec:dilemma}


Self-reflection mechanisms, a pivotal strategy for enhancing model performance, were initially applied to LLMs to bolster their complex reasoning capabilities. This approach has since demonstrated promising outcomes when extended to multimodal contexts. As illustrated in the left part of \cref{fig: motivation}, the representative VL-Rethinker method demonstrates substantial performance improvements across various image understanding benchmarks. However, despite their operation within multimodal frameworks, the reflection processes of such mechanisms remain purely text-based. They do not incorporate re-evaluation of visual inputs, thereby classifying them as text-driven reflection mechanisms.

As shown in the middle part of \cref{fig: motivation}, applying these text-based reflection mechanisms to long-form video understanding tasks paradoxically leads to a decline in model performance. To eliminate the possibility that this degradation stems from VL-Rethinker’s lack of video training, we developed a comparable model equipped with a purely text-based reflection mechanism and trained it on video data. Details are provided in Appendix \cref{app: training reflection}. Our experiments revealed that this model also failed to improve performance in long-form video understanding tasks.

We attribute this phenomenon to two primary factors: (1) \textit{Compared to static images, long-form videos encompass significantly richer and more dynamic visual content. Thus, re-evaluating only text-based representations is inadequate for correcting reasoning errors, necessitating a visual re-evaluation process.} (2) \textit{Purely text-based reflection mechanisms inherently lack cross-modal interaction capabilities, preventing full integration of visual information during reflection and thereby limiting improvements in MLLMs for long-form video understanding.}

\subsection{Necessity of Integrating Visual Rethinking}
\label{subsec:necessity}

To substantiate the claim advanced in \cref{subsec:dilemma}, specifically that re-evaluating visual information can improve MLLMs’ accuracy in long-form video comprehension, we conducted a preliminary validation experiment. In datasets such as NExT-GQA \citep{xiao2024can}, ReXTime \citep{chen2024rextime}, and CG-Bench \citep{chen2024cg}, the critical video segments necessary for answering each question, representing only a small fraction of the total footage, are annotated. Using these datasets, we first allowed MLLMs to perform initial reasoning based on the raw video frames and questions. Subsequently, we provided the models with the annotated key segments and instructed them to re-evaluate the problem, integrating their previous reasoning with the newly supplied visual cues to generate the final answers.

As illustrated in the right part of \cref{fig: motivation}, introducing a visual rethinking process during the reflection phase yields an average accuracy gain of approximately 7.3\% on the NExT-GQA, ReXTime, and CG-Bench datasets. In contrast, reflection based solely on textual information results in negligible improvement. These results substantiate two key insights: (1) \textit{In the domain of video understanding, visual reflection plays a far more crucial role than textual reflection}; and (2) \textit{when MLLMs effectively revisit key visual cues during this rethinking stage, their accuracy in long-form video comprehension tasks can be substantially enhanced.}    
\section{Methodology}
\label{sec:method}

In this section, we present REVISOR (REflective VIsual Segment Oriented Reasoning), a novel two-stage reasoning framework designed to enhance long-form video understanding. REVISOR leverages a tool-augmented multimodal reflection mechanism that enables MLLMs to collaboratively construct reflection processes across textual and visual modalities, enhancing their reasoning capability.

\begin{figure*}[t]
  \centering
   \includegraphics[width=0.95\linewidth]{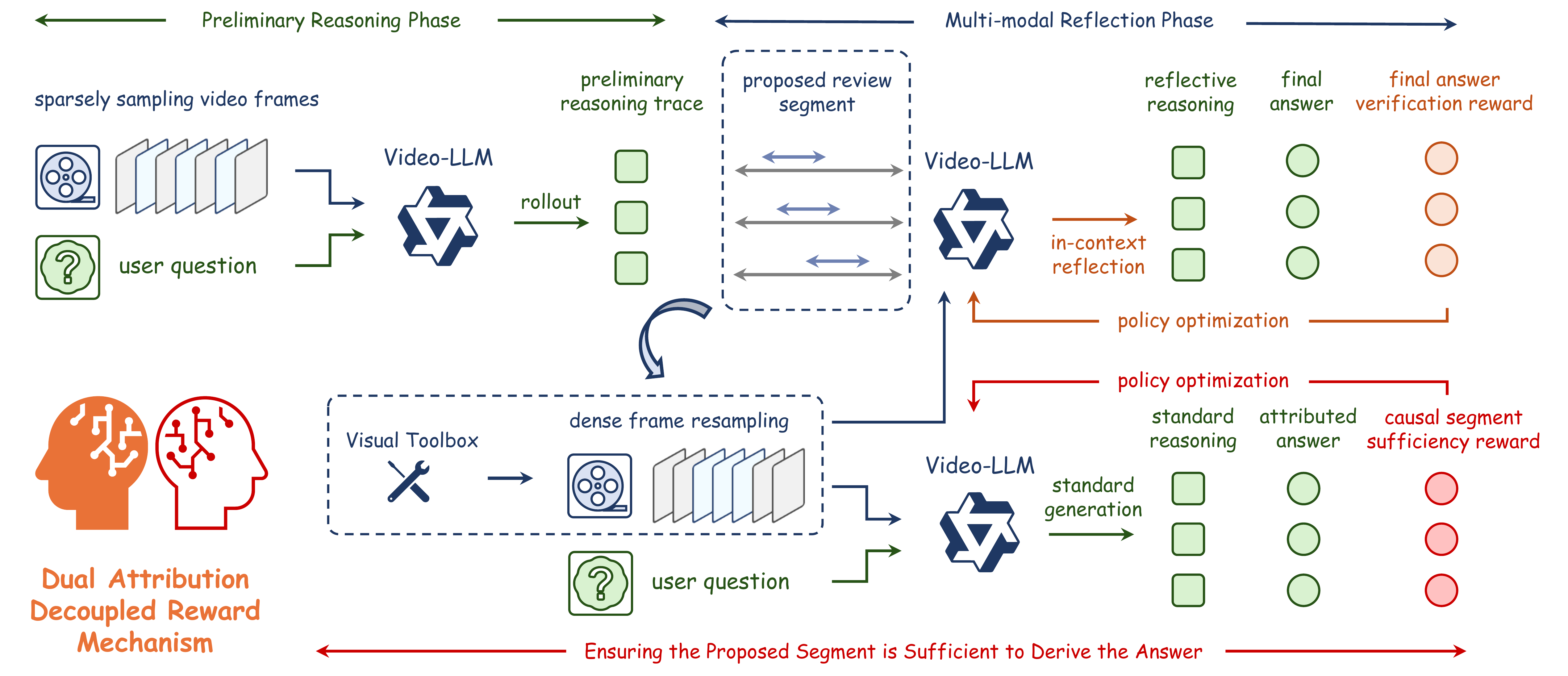}
   \caption{Overview of the Dual-Attribution Decoupled Reward Mechanism (DADR). Final Answer Verification Reward (top) is derived from verifying the correctness of the model's synthesized final answer, directly targeting the accuracy objective of the reflective stage. Conversely, Causal Segment Sufficiency Reward (bottom) is granted upon verifying an attribution answer derived exclusively from reviewed video segments, thereby guiding the model to identify and utilize segments highly pertinent to the user query.}
   \label{fig: method dadr}
\end{figure*}

\subsection{Tool-Augmented Multimodal Reflection}
\label{subsec:framework}

The core concept of REVISOR involves a two-stage reasoning framework: (1) an initial inference stage that generates a preliminary reasoning trace and identifies key moments for further examination, and (2) a reflective reasoning stage that integrates the initial analysis with newly sampled, fine-grained visual evidence to yield a refined final answer. An overview of this information flow is illustrated in \cref{fig: method revisor}.

\vspace{0.15cm}
\noindent \textbf{Stage 1: Initial Inference and Visual Review Proposal.} The process begins with a video $V$ and a user-posed question $Q$. To mitigate the computational cost and context-length limitations of the MLLM, we first perform a sparse, uniform sampling of frames from the video, producing an initial frame set $F_{init} = {f_1, f_2, \dots, f_N}$, where $N$ denotes the number of sampled frames.

The MLLM, denoted as $\mathcal{M}$, takes as input a question $Q$ and the initial set of frames $F_{init}$. The model is prompted to engage in a chain-of-thought reasoning process to generate a preliminary reasoning trace. Importantly, beyond producing the reasoning content, the model is instructed to identify and output a temporal segment $S$ that it considers most relevant or uncertain with respect to its conclusion. This output is represented as a structured tuple containing the initial reasoning trace $T_{init}$ and the proposed review segment $S$. We formalize this stage as:
\begin{equation}
    (T_{init}, S) = \mathcal{M}_{infer}(Q, F_{init}),
\end{equation}
where $\mathcal{M}_{infer}$ denotes the MLLM operating in its initial inference mode. The segment $S = [t_{start}, t_{end}]$ specifies the start and end timestamps of the video interval that warrants closer examination. The model’s ability to propose $S$ emerges naturally from its reasoning process $T_{init}$, through which it articulates which portions of the video were most pivotal or ambiguous.

\vspace{0.15cm}
\noindent \textbf{Visual Toolbox Call: Retrieve Review Segment Frames.} Upon receiving the proposed segment $S$ from the MLLM, the REVISOR framework engages the Visual Toolbox, denoted as $\mathcal{T}$. The role of $\mathcal{T}$ is to perform a targeted, high-density re-sampling of frames from the original video $V$, restricted to the temporal window specified by $S$.

For a segment $S = [t_{start}, t_{end}]$, the toolbox generates a denser frame sequence, formalized as:
\begin{equation}
    F_{review} = \text{SampleDense}(V, [t_{start}, t_{end}]),
\end{equation}
where $\text{SampleDense}(\cdot)$ denotes a sampling function that extracts frames at a higher temporal resolution (e.g., increased FPS) compared to the initial sparse sampling. This tool-assisted process offloads the computational and procedural burden of locating fine-grained visual details, enabling the MLLM to conduct a focused examination of critical moments without processing the entire video at full resolution.

\vspace{0.15cm}
\noindent \textbf{Stage 2: Reflective Reasoning and Answer Refinement.} The final stage centers on reflection and self-correction. The MLLM is re-invoked, now operating within a richer and more structured contextual frame. The inputs to this reflective stage are: (1) the original question $Q$, (2) the initial reasoning trace $T_{init}$, and (3) the newly acquired, densely sampled visual evidence $F_{review}$.

By providing the model with access to its own initial reasoning $T_{init}$, the REVISOR framework enables a form of in-context reflection, prompting the model to reassess its preliminary conclusions in light of the enhanced visual data from $F_{review}$. This process allows the MLLM to validate its earlier hypotheses, resolve ambiguities identified in Stage 1, or correct prior misinterpretations. The generation of the refined reasoning trace $T_{refine}$ and final answer $A_{final}$ is formalized as:
\begin{equation}
    (T_{refine}, A_{final}) = \mathcal{M}_{reflect}(Q, T_{init}, F_{review}),
\end{equation}
where $\mathcal{M}_{reflect}$ denotes the MLLM operating in its reflection mode. This reflective mechanism mirrors human expert analysis, where an initial overview is followed by a focused examination of critical evidence before forming a conclusion. By structuring the MLLM’s reasoning in this iterative manner, REVISOR enhances both the accuracy and reliability of responses in complex video understanding tasks.

\subsection{Dual Attribution Decoupled Reward}
\label{subsec:reward}

In standard practice, reward signals based on the verification of the final answer are integrated with reinforcement learning to train the aforementioned tool-augmented MLLM in an end-to-end manner. Specifically, the GRPO algorithm is commonly employed to maximize the expected total reward, thereby stabilizing the training process and encouraging exploration within the complex video–temporal action space. The policy $\pi_{\theta}$ parameterized by $\theta$ generates a reasoning trajectory $\tau = (T_{init}, S, T_{refine}, A_{final})$ based on the input $(Q, V)$. The overall objective is to maximize the expected cumulative reward $J(\theta)$:
\begin{equation}
    J(\theta) = \mathbb{E}_{(Q, V) \sim \mathcal{D}, \tau \sim \pi_{\theta} | (Q, V)} [\mathcal{R}(\tau)].
\end{equation}

However, as shown in \cref{tab:reward_ablation}, training the REVISOR framework using the GRPO algorithm with final-answer verification as the sole reward function leads to decreased performance on long-form video understanding tasks compared to the base model. This limitation primarily arises from the complex nature of the REVISOR reasoning process, which consists of three components: the initial reasoning content, the video review segmentation interval, and the reflective reasoning phase. Consequently, even when the model correctly outputs the video review segments, it receives insufficient positive feedback; conversely, incorrect segment predictions fail to incur adequate penalties. This imbalance prevents the model from effectively learning to identify key video segments during reinforcement learning (see \cref{subsec:segment-accuracy} for supporting evidence).

To address this issue, we decouple the reward for video review segment localization from the general reward signal and propose a Dual-Attribution Decoupled Reward Mechanism (DADR), as shown in \cref{fig: method dadr}. This mechanism integrates the reward for the final output with an additional causal reward, ensuring that the proposed segment $S$ is sufficient for deriving the correct conclusion. The total reward $\mathcal{R}(\tau)$ is defined as a weighted sum of two distinct components: the Final Answer Verification Reward ($R_{final}$) and the Causal Segment Sufficiency Reward ($R_{causal}$):
\begin{equation}
\mathcal{R}(\tau) = \lambda_{1} R_{final} + \lambda_{2} R_{causal},
\end{equation}
where $\lambda_{1}$ and $\lambda_{2}$ are hyperparameters controlling the balance of the rewards. The term $R_{final}$ is a standard task-specific reward that validates the correctness of the final refined answer $A_{final}$ against the ground truth answer $A^*$.

\definecolor{myred}{HTML}{CB0000}
\begin{table*}[t]
\small
\centering
\caption{Evaluation results of the four long-form video understanding benchmarks. $^{\ast}$ indicates models trained with the text-based reflection mechanism using the same dataset as ours. $^{\star}$ represents our reproduction. $\uparrow$ highlights the superior performance of the REVISOR framework relative to the base model Qwen2.5-VL-7B.}
\vspace{-0.2cm}
\label{tab:longvideo}
\begin{tabular}{@{}ccccc|ccc@{}}
\toprule[1pt]
\toprule
\multicolumn{1}{c|}{}                                 & \multicolumn{1}{c|}{}                                & \multicolumn{1}{c|}{}                & \multicolumn{2}{c|}{\textbf{VideoMME}} &                                           &                                 &                                    \\ \cmidrule(lr){4-5}
\multicolumn{1}{c|}{\multirow{-2}{*}{\textbf{Model}}} & \multicolumn{1}{c|}{\multirow{-2}{*}{\textbf{Model Size}}} & \multicolumn{1}{c|}{\multirow{-2}{*}{\textbf{Video Tokens}}} & \textbf{Long}    & \textbf{Overall}   & \multirow{-2}{*}{\textbf{LongVideoBench}} & \multirow{-2}{*}{\textbf{MLVU}} & \multirow{-2}{*}{\textbf{LVBench}} \\ \midrule
\multicolumn{1}{c|}{Gemini-1.5-Pro\citep{team2024gemini}}    & \multicolumn{1}{c|}{-}               & \multicolumn{1}{c|}{-}                                             & 67.4             & 75                 & 64                                        & -                               & -                                  \\
\multicolumn{1}{c|}{GPT4o\citep{hurst2024gpt}}      & \multicolumn{1}{c|}{-}                      & \multicolumn{1}{c|}{-}                                             & 65.3             & 71.9               & 66.7                                      & 64.6                            & -                                  \\ \midrule
\multicolumn{1}{c|}{ShareGPT4Video\citep{chen2024sharegpt4video}}       & \multicolumn{1}{c|}{8B}                    & \multicolumn{1}{c|}{-}                                          & 37.9             & 43.6               & 39.7                                         &          46.4                   & -                                  \\

\multicolumn{1}{c|}{Video-LLaVA\citep{lin2024video}}       & \multicolumn{1}{c|}{7B}                    & \multicolumn{1}{c|}{-}                                          & 38.1             & 40.4               & 39.1                                         &          47.3                   & -                                  \\
\multicolumn{1}{c|}{LongVA\citep{zhang2024long}}       & \multicolumn{1}{c|}{7B}                    & \multicolumn{1}{c|}{224K}                                          & 47.6             & 54.3               & -                                         & 56.3                            & -                                  \\
\multicolumn{1}{c|}{LongVU\citep{shen2024longvu}}     & \multicolumn{1}{c|}{7B}                     & \multicolumn{1}{c|}{8K}                                            & 59.5             & 60.6               & -                                         & 65.4                            & -                                  \\
\multicolumn{1}{c|}{Vamba\citep{ren2025vamba}}     & \multicolumn{1}{c|}{10B}                     & \multicolumn{1}{c|}{-}                                            & -             & 57.8               & 55.9                                         & 65.9                            & 42.1                                  \\
\multicolumn{1}{c|}{LLaVA-OneVision\citep{li2024llava}}      & \multicolumn{1}{c|}{7B}           & \multicolumn{1}{c|}{6K}                                            & -                & 46.7               & 56.4                                      & 64.7                            & -                                  \\
\multicolumn{1}{c|}{Hour-LLaVA\citep{lin2025unleashing}}     & \multicolumn{1}{c|}{7B}                    & \multicolumn{1}{c|}{-}                                              & 55               & 63.6               & 60.4                                      & -                               & 45.6                               \\
\multicolumn{1}{c|}{VideoChat-Flash\citep{li2025videochatflashhierarchicalcompressionlongcontext}}         & \multicolumn{1}{c|}{7B}                  & \multicolumn{1}{c|}{8K}                                              &      55.4        &  65.3              &           -                           &          74.7                  &  48.2                                 \\ 

\multicolumn{1}{c|}{NVILA\citep{liu2025nvila}}         & \multicolumn{1}{c|}{8B}                  & \multicolumn{1}{c|}{8K}                                              & 54.8             & 64.2               & 70.1                                      & 57.7                            & -                                  \\ \midrule
\multicolumn{1}{c|}{Open-o3-Video\citep{meng2025open}}       & \multicolumn{1}{c|}{7B}               & \multicolumn{1}{c|}{2K}                                            & 54.9             & 63.6               & -                                         & -                               & -                                  \\
\multicolumn{1}{c|}{Video-MTR\citep{xie2025video}}       & \multicolumn{1}{c|}{7B}               & \multicolumn{1}{c|}{4K}                                            & 51.0             & 59.0               & -                                         & 48.4                               & -                                  \\

\multicolumn{1}{c|}{Video-R1\citep{feng2025video}}         & \multicolumn{1}{c|}{7B}               & \multicolumn{1}{c|}{8K}                                            & -                & 61.4               & -                                         & -                               & -                                  \\ 

\multicolumn{1}{c|}{LongVILA-R1\citep{chen2025scaling}}         & \multicolumn{1}{c|}{7B}               & \multicolumn{1}{c|}{-}                                            & 55.2                & 65.1               & 58                                         & -                               & -                                  \\ 
\midrule
\multicolumn{1}{c|}{VL-Rethinker\citep{wang2025vl}}         & \multicolumn{1}{c|}{7B}              & \multicolumn{1}{c|}{8K}                                            & 51.9             & 62.1               & 56.4                                      & 63.2                            & 37.2                               \\
\multicolumn{1}{c|}{Qwen2.5-VL$^{\ast}$\citep{bai2025qwen2}}       & \multicolumn{1}{c|}{7B}               & \multicolumn{1}{c|}{8K}                                            & 53.2             & 63.4               & 57.4                                      & 69                              & 39.2                               \\ 
\multicolumn{1}{c|}{Qwen2.5-VL$^{\star} $\citep{bai2025qwen2}}       & \multicolumn{1}{c|}{7B}               & \multicolumn{1}{c|}{8K}                                            & 53.4             & 64.3               & 56.5                                      & 67.3                            & 40.2                               \\
\rowcolor[HTML]{D8F4FF} 
\multicolumn{1}{c|}{Ours}             & \multicolumn{1}{c|}{7B}        & \multicolumn{1}{c|}{8K}                                            & \textbf{56.2} \textcolor{myred}{$\uparrow$2.8} & \textbf{65.7} \textcolor{myred}{$\uparrow$1.4} & \textbf{57.5} \textcolor{myred}{$\uparrow$1.0} & \textbf{69.8} \textcolor{myred}{$\uparrow$2.5} & \textbf{42} \textcolor{myred}{$\uparrow$1.8} \\ \bottomrule
\bottomrule[1pt]
\end{tabular}
\end{table*}


\vspace{0.15cm}
\noindent \textbf{Causal Segment Sufficiency Reward.} The Causal Segment Sufficiency Reward (CSSR) is designed to enforce the quality of the proposed review segment $S$ (generated in Stage 1). The CSSR provides a positive signal only if the model is capable of deriving the correct answer $A^*$ solely based on the question $Q$ and the densely sampled visual evidence $F_{review}$ extracted from $S$. This implicitly encourages the MLLM to select segments that are truly causal and sufficient for the task, preventing the selection of irrelevant or overly long segments.

To formalize this, we define an attribution-based answer $\hat{A}$ using the same MLLM $\mathcal{M}$, but conditioned solely on the original question $Q$ and the fine-grained visual evidence $F_{review}$. This step serves as a crucial sufficiency test for the proposed segment $S$. Let $\mathcal{M}_{suff}$ denote this sufficiency prediction mode:
\begin{equation}
    \hat{A} = \mathcal{M}_{suff}(Q, F_{review}).
\end{equation}
The Causal Segment Sufficiency Reward is then defined based on the correctness of this attribution-based prediction:
\begin{equation}
    R_{causal} = \mathbb{I}(\hat{A} = A^{*}).
\end{equation}
Incorporating $R_{causal}$ into the GRPO optimization loop guides the REVISOR framework toward policies $\pi_{\theta}$ that not only yield accurate final answers but also robustly identify the critical temporal evidence required for those answers in the initial inference stage.

\section{Experiments}
\label{sec:experiments}

We first detail our experimental setup in \cref{subapp: experiment setup}. The performance of the REVISOR framework for long-form video understanding tasks is then presented in \cref{main_results}. Subsequently, \cref{ablation_study} conducts an ablation study to evaluate the contribution of REVISOR's various components. 
Please refer to Appendix \cref{app: case study} for the Case Study of REVISOR.

\subsection{Experimental Setup}
\label{expertimental_setup}

\noindent\textbf{Training Details.} We adopt Qwen2.5-VL-7B \citep{bai2025qwen2} as our base model. The training is conducted using the verl framework \citep{sheng2025hybridflow}, which we further extend to support the training of REVISOR. Our training process consists of a single-stage reinforcement learning phase following DAPO. \cite{yu2025dapo}. 

\noindent\textbf{Dataset Construction.} We use a total of 25K training samples collected from STAR \citep{li2025star}, PerceptionTest \citep{patraucean2023perception}, NExT-QA \citep{xiao2021next}, CLEVRER \citep{yi2019clevrer}, LLaVA-Video-178K \citep{zhang2024video}, TimeRFT \citep{wang2025time}, CG-Bench \citep{chen2024cg}, and ReXTime \citep{chen2024rextime}. Detailed information regarding the dataset selection procedure is available in Appendix \cref{subapp: dataset}.

\noindent\textbf{Hyperparameter Selection.} In our experiments, we set $\lambda_{1} = 0.6$ and $\lambda_{2} = 0.3$. We adopt AdamW \citep{loshchilov2017decoupled} as the optimizer with a learning rate of $1\times10^{-6}$ and a batch size of $32$. The number of rollouts is set to $8$. During both training and evaluation, the total number of video tokens in the input is limited to a maximum of $8192$. For more details, please refer to Appendix \cref{subapp: experiment setup}

\subsection{Main Results}
\label{main_results}

\textbf{Long-Form Video Understanding Task.} We evaluate our method on four widely used long-form video benchmarks: VideoMME \citep{fu2025video}, LongVideoBench \citep{wu2024longvideobench}, MLVU \citep{zhou2025mlvu}, and LVBench \citep{wang2025lvbench}. As shown in \cref{tab:longvideo}, with an input of 8K video tokens, REVISOR achieves 65.7\% on VideoMME, outperforming the base model Qwen2.5-VL-7B by 1.4\%. Notably, REVISOR outperforms the base model by 2.8\% on the long subset of VideoMME and by 2.5\% on MLVU, which contains videos up to 120 minutes. This indicates that as video duration increases, accurately reviewing the relevant video content becomes increasingly crucial. 

Compared with the latest text-only reasoning approach Video-R1 \citep{feng2025video}, REVISOR achieves a 4.3\% improvement on VideoMME, demonstrating that the integration of the DADR mechanism provides a clear advantage over pure text-based reasoning. Furthermore, compared with text-based reflection methods such as VL-Rethinker and our own video-data-trained, text-only reflection method, REVISOR shows gains of 3.6\% and 2.3\%, respectively, highlighting the necessity of Visual Rethinking within REVISOR.

\vspace{-0.1cm}
\begin{table}[h]
\small
\centering
\caption{Evaluation results for the temporal video grounding task, including the Charades-STA and NExT-GQA benchmarks. \textbf{Bold} fonts highlight the best performance.}
\vspace{-0.2cm}
\label{tab:temporal_grounding}
\resizebox{\linewidth}{!}{
\begin{tabular}{@{}c|cc|cc@{}}
\toprule[1pt]
\toprule
                                 & \multicolumn{2}{c|}{\textbf{Charades-STA}}                       & \multicolumn{2}{c}{\textbf{NExT-GQA}}                            \\ \cmidrule(l){2-5} 
\multirow{-2}{*}{\textbf{Model}} & \textbf{R@0.7} & \textbf{mIoU} & \textbf{R@0.7} & \textbf{mIoU} \\ \midrule
Qwen2.5-VL-7B\citep{bai2025qwen2}                    & 15.5          & 36.9         & 7.5           & 20.9         \\
VTimeLLM\citep{huang2024vtimellm}                         & 14.7           & 34.6          & 9.7            & 24.4          \\
iMOVE\citep{li2025imove}                            & 26.1           & 47.3          & -              & -             \\
TimeChat\citep{ren2024timechat}                         & 13.4           & -             & 6.2            & 20.6          \\
VideoChat-TPO\citep{yan2025task}                    & 18.4           & 38.2          & 8.2            & 27.7          \\
TVG-R1\citep{chen2025datasets}                           & 23.9          & 46.7          & 10             & 29.3          \\
\midrule
\rowcolor[HTML]{D8F4FF} 
Ours                             & \textbf{31.8}  & \textbf{51.4} & \textbf{11.9} & \textbf{33.2} \\ \bottomrule
\bottomrule[1pt]
\end{tabular}
}
\end{table}

\noindent\textbf{Temporal Video Grounding Task.} As shown in \cref{tab:temporal_grounding}, REVISOR achieves 51.4\% mIoU on Charades-STA \citep{gao2017tall}, surpassing the prior SFT-based SOTA method iMOVE \citep{li2025imove} by 4.1\% mIoU. Moreover, it outperforms the RL-based temporal grounding model TVG-R1 \citep{chen2025datasets} by 4.7\% mIoU on Charades-STA and 3.9\% mIoU on NExT-GQA \citep{xiao2024can}, showing its superior capability in temporal video grounding.

\begin{table}[h]
\small
\centering
\caption{Ablation study of the Dual Attribution Decoupled Reward mechanism. V-MME, LongVB, LV, and NExT-G represent VideoMME, LongVideoBench, LVBench, and NExT-GQA, respectively. \textbf{Bold} fonts highlight the best performance. The row marked in gray represents our base model.}
\vspace{-0.2cm}
\label{tab:reward_ablation}
\begin{tabular}{@{}cc|ccccc@{}}
\toprule[1pt]
\toprule
$\boldsymbol{\lambda_1}$ & $\boldsymbol{\lambda_2}$ & \textbf{V-MME} & \textbf{LongVB} & \textbf{LV} & \textbf{MLVU} & \textbf{NExT-G} \\ 
\midrule
\rowcolor[HTML]{EFEFEF} - & - & 64.3 & 56.5 & 40.2 & 67.3 & 20.87 \\ 
\midrule
0.3 & 0.6 & 64.0 & 56.0 & 41.1 & 68.7 & 33.9 \\
0.6 & 0.0 & 62.2 & 54.0 & 40.8 & 68.3 & 32.1 \\
\rowcolor[HTML]{D8F4FF} 0.6 & 0.3 & \textbf{65.7} & \textbf{57.5} & \textbf{42.0} & \textbf{69.8} & \textbf{33.2} \\ 
\bottomrule
\bottomrule[1pt]
\end{tabular}
\end{table}

\vspace{-0.2cm}
\subsection{Ablation Study}
\label{ablation_study}

\textbf{Dual Attribution Decoupled Reward.} We perform an in-depth investigation of the DADR mechanism. As shown in \cref{tab:reward_ablation}, setting $\lambda_{2}$ to 0 (i.e., using only the Final Answer Verification Reward) results in a drop of REVISOR's Video-MME score from 65.7\% to 62.2\%, which is below the 64.3\% achieved by the base model Qwen2.5-VL-7B. This result suggests that, without the CSSR component, the model fails to learn how to locate the correct review segment $S$ from sparse reward signals. Please refer to Appendix \cref{subapp: additional ablation} for more ablation study results.

When the value of $\lambda_{2}$ exceeds that of $\lambda_{1}$, the model’s temporal video grounding ability improves, but its long-form video understanding performance declines (e.g., MLVU drops from 69.8\% to 68.7\%). This occurs because the model becomes overly focused on locating the correct review segment $S$, while neglecting how to utilize $S$ to derive the correct answer. Therefore, we set the value of $\lambda_{1}$ to be greater than that of $\lambda_{2}$ to encourage the model not only to accurately identify the correct review segment $S$ but also to leverage it to enhance its reasoning capability. 

\vspace{-0.1cm}

\begin{table}[h]
\small
\centering
\caption{Ablation study of the training data composition. SVQ, TRF, RXT, and CGB represent short video QA, TimeRFT, ReXTime, and CG-Bench, respectively. V-MME, LVB, LV, and NExT-G represent VideoMME, LongVideoBench, LVBench, and NExT-GQA, respectively. \textbf{Bold} fonts highlight the best performance. The row marked in gray represents our base model.}
\vspace{-0.2cm}
\label{tab:dataset_ablation}
\setlength{\tabcolsep}{2pt} 
\begin{tabular}{@{}cccc|ccccc@{}}
\toprule[1pt]
\toprule
\textbf{SVQ} & \textbf{TRF} & \textbf{RXT} & \textbf{CGB} & \textbf{V-MME} & \textbf{LVB} & \textbf{LV} & \textbf{MLVU} & \textbf{NExT-G} \\ 
\midrule
\rowcolor[HTML]{EFEFEF} - & -& -& -& 64.3 & 56.5 & 40.2 & 67.3 & 20.9 \\
\midrule
\ding{51} & & & & 62.6 & 52.7 & 38.6 & 66.7 & 25.0 \\
\ding{51} & \ding{51} & & & 63.8 & 56.8 & 40.3 & 67.6 & 22.6 \\
\ding{51} & \ding{51} & \ding{51} & & 64.0 & 57.1 & 40.5 & 68.4 & 29.5 \\
\rowcolor[HTML]{D8F4FF} \ding{51} & \ding{51} & \ding{51} & \ding{51} & \textbf{65.7} & \textbf{57.5} & \textbf{42.0} & \textbf{69.8} & \textbf{33.2} \\
\bottomrule
\bottomrule[1pt]
\end{tabular}
\end{table}

\noindent\textbf{Data Composition.} As shown in \cref{tab:dataset_ablation}, REVISOR's long-form video understanding capability improves with the inclusion of more diverse datasets. Specifically, its performance on MLVU increases from 66.7\% to 69.8\%. Notably, even when trained exclusively on short video QA datasets, REVISOR's mIoU on NExT-GQA rises from 20.9\% to 25\%. This result demonstrates that, when combined with the DADR mechanism, our framework can effectively translate its understanding into the accurate localization of the correct review segment $S$.

\section{Extended Analysis}
\label{sec:discussion}

This section details comprehensive analytical experiments validating two key factors crucial for the REVISOR framework's success: (1) For long-form video understanding tasks, the re-examination of visual information proves significantly more critical than that of textual inference processes. (2) The DADR mechanism effectively enhances the MLLM's precision in retrieving critical visual information.


\subsection{Importance of Visual Rethinking}
\label{subsec:discuss-visual}

\cref{sec:motivation} posited and preliminarily demonstrated that the limitations of purely text-based reflection in long-form video understanding arise from the inherently richer and more dynamic nature of long-form video inputs compared to static images. Reflection based solely on textual information is insufficient to correct reasoning errors, necessitating the integration of a dedicated visual reflection process. In this section, we provide further indirect evidence supporting this claim through an analysis of REVISOR's output.

\begin{figure}[h]
  \centering
   \includegraphics[width=0.99\linewidth]{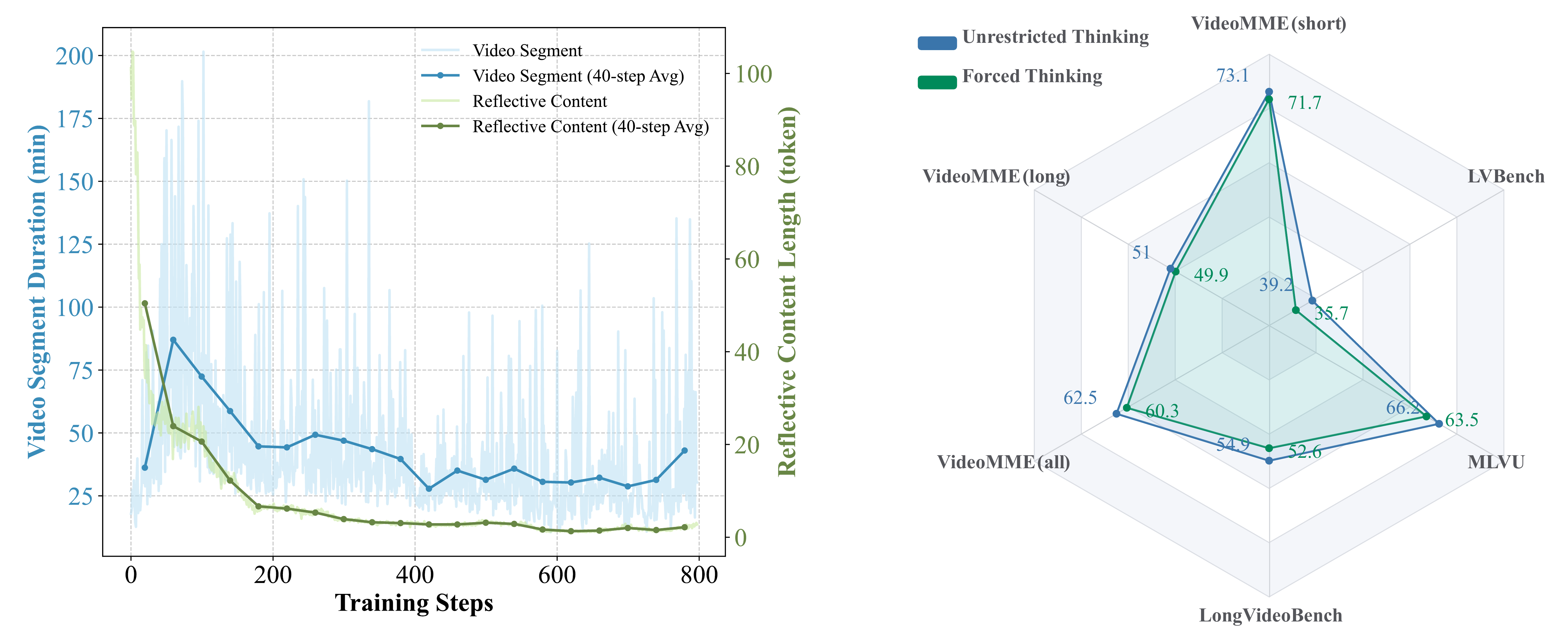}
   \vspace{-0.2cm}
   \caption{The superior efficacy of visual reflection over textual reflection in long-form video understanding. The left panel demonstrates that the length of the generated textual reflection consistently decreases throughout training. The right panel further indicates that forcing the model to perform longer textual reflection actually leads to a degradation in model performance.}
   \label{fig: discuss 01}
\end{figure}

Left part of \cref{fig: discuss 01} depicts the evolution of textual reasoning length and video review length during REVISOR’s reflection stage throughout training. As training progresses, the length of textual reasoning steadily decreases. This trend clearly indicates that, through interaction with the environment during reinforcement learning, the model gradually learns that textual reflection plays a relatively minor role in long-form video understanding tasks. In contrast, the length of the reviewed video segments first increases and then decreases. We attribute this to the model initially expanding its search range for critical video segments to ensure the inclusion of highly relevant content, and later learning to discard redundant portions. This refinement enables the model to precisely locate the minimal video segments sufficient for answering the given question. 

Building on these findings, we conducted further comparative experiments on the length of the text generated during the reflection process. By encouraging the model to engage in deeper deliberation through the system prompt, we made the model output more extensive textual reasoning during the reflection stage. However, as shown in the right part of \cref{fig: discuss 01}, this intervention paradoxically led to a decline in the model's performance on the video understanding task. This result further substantiates that pure textual reflection is not critical for video understanding tasks; instead, generating more ineffectual reasoning content can have a negative impact on performance.

\subsection{Accuracy of Visual Information Retrieval}
\label{subsec:segment-accuracy}

In \cref{subsec:reward}, we propose a DADR mechanism to enable more accurate revisiting of key video segments during the reflection phase. To assess its efficacy, we conduct targeted experiments on the video segments REVISOR processes within the visual reflection stage.

We aim to examine the effect of incorporating the DADR mechanism during training on the accuracy of retrospective video segment extraction. To this end, we conducted an experiment in which models trained with and without the DADR mechanism were required to extract key video segments necessary for answering specific questions. These extracted segments were subsequently provided to the original Qwen2.5-VL-7B model, which generated responses based solely on the given segments. As the model architecture and parameters remained identical, the correctness of the responses depended entirely on the relevance of the retrieved video segments to the corresponding questions. Thus, by comparing the answer accuracy of Qwen2.5-VL-7B under both conditions, we can indirectly assess the precision of the video segment extraction process.

\begin{figure}[h]
  \centering
   \includegraphics[width=0.99\linewidth]{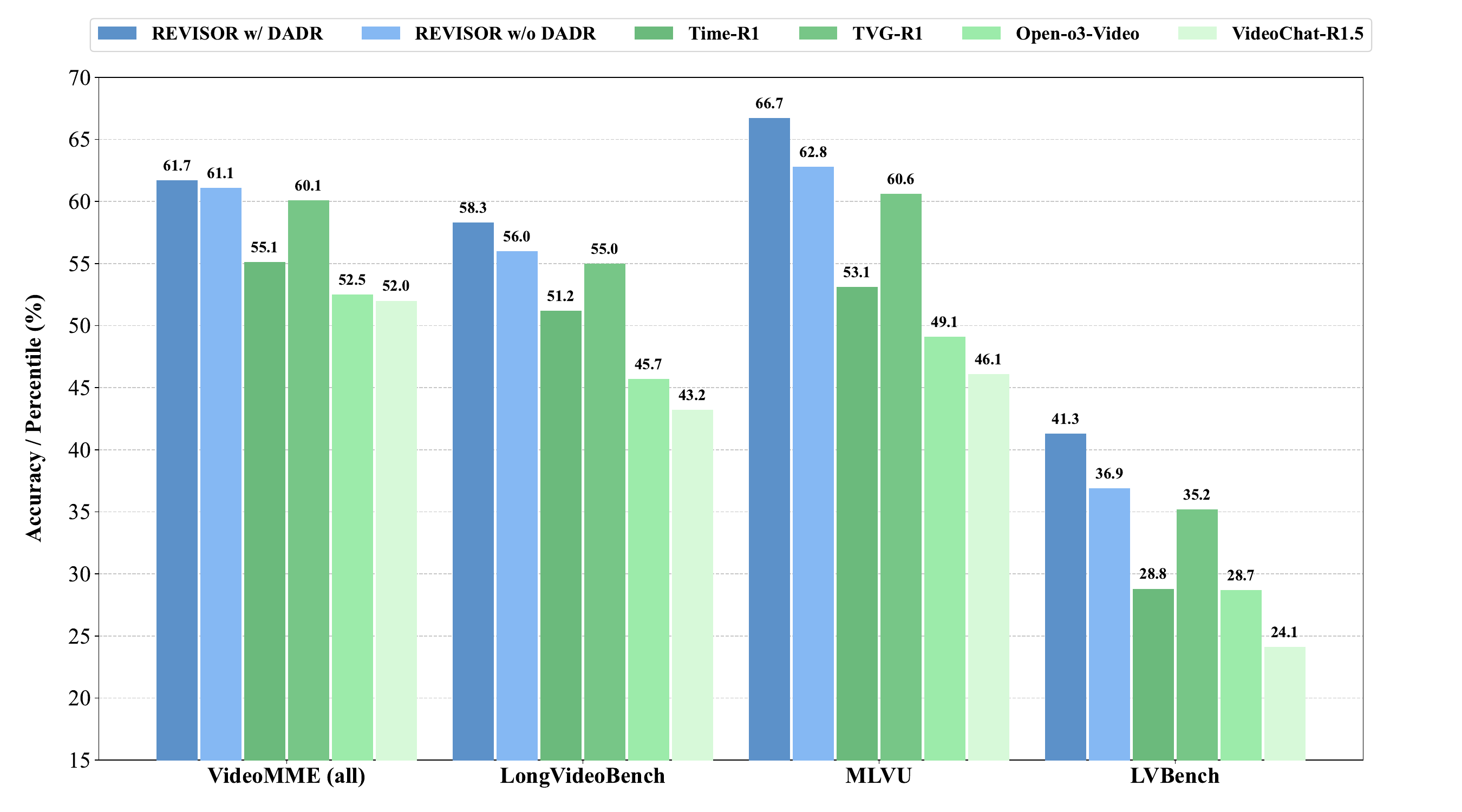}
   \vspace{-0.2cm}
   \caption{Comparative accuracy of key moment review across different methods. Methods based on the REVISOR framework and its variants are highlighted in blue, while different Temporal Grounding baselines are represented in green.}
   \label{fig: discuss 02}
\end{figure}

As shown in \cref{fig: discuss 02}, when the DADR mechanism is applied during training, the video segments revisited by the model during the reflection phase are generally more accurate. This clearly demonstrates that the DADR mechanism enables MLLMs to learn to recall key video segments during the reinforcement learning process. Furthermore, we conducted a comparative analysis between the DADR mechanism and the four temporal grounding methods Time-R1 \citep{wang2025time}, TVG-R1 \citep{chen2025datasets}, Open-o3-Video \citep{meng2025open}, and VideoChat-R1.5 \citep{yan2025videochat}, and found that DADR still achieves the highest accuracy in recalling relevant video segments.
\section{Conclusion}
\label{sec:conclusion}

Text-based reflection fails to capture the complexity of long-form video reasoning, which depends critically on rich and dynamic visual information. We propose REVISOR, a tool-augmented multimodal reflection framework that revisits essential visual content to correct errors arising in the initial reasoning process. To train REVISOR to precisely localize the video segments most relevant to each query, we incorporate a Dual Attribution Decoupled Reward (DADR) mechanism into GRPO. Extensive experiments across four long-form video understanding benchmarks show that REVISOR consistently delivers significant performance gains. For related work, please refer to Appendix \cref{sec:related work}.

{
    \small
    \bibliographystyle{ieeenat_fullname}
    \bibliography{main}
}

\clearpage
\setcounter{page}{1}
\maketitlesupplementary

\section{Related Work}
\label{sec:related work}

\textbf{Long-Form Video Understanding.} In long-form video understanding, the visual inputs are far more complex than those in image-based tasks. Consequently, identifying and distilling only the question-relevant information from large amounts of redundant visual content becomes crucial for improving model performance. Existing approaches can be grouped into three categories: external-model augmentation, agentic methods, and model-internalized selection.

External-model augmentation techniques \citep{man2025adacm2understandingextremelylongterm, shen2024longvu, liang2024keyvideollmlargescalevideokeyframe}, rely on vision–text similarity models like CLIP for keyframe selection. While efficient, these methods perform static, one-shot selection, limiting their adaptability to complex or evolving queries. Agentic approaches \citep{wang2025videochata1thinkinglongvideos, liu2025videomindchainofloraagentlong} address this limitation through iterative frameworks that dynamically refine frame selection. For example, VideoAgent \citep{wang2024videoagentlongformvideounderstanding} employs an LLM as a central agent in a state–action–observation loop, whereas VideoTree \citep{wang2025videotreeadaptivetreebasedvideo} adopts a hierarchical tree structure to perform coarse-to-fine search, directing computation toward the most relevant segments. More recent model-internalized selection methods \citep{tong2025thinkingvideovideogeneration} embed frame selection directly into the model’s reasoning process. GenS \citep{yao2025generativeframesamplerlong} trains a lightweight generative sampler to identify query-relevant frames end to end, though the selection and QA stages remain decoupled.

Our proposed REVISOR framework, after reinforcement learning training, can autonomously identify and explore visual content that requires additional careful review during the inference stage. Unlike previous methods that are limited to statically searching for important video segments based solely on the question itself, our approach fully integrates with the reasoning capabilities of MLLMs. This allows it to engage in deeper deliberation and, through comprehensive interaction with its ongoing reasoning, more precisely pinpoint critical visual information, thereby significantly enhancing the model's reasoning ability.

\vspace{0.15cm}
\noindent \textbf{Self-Reflection Mechanism.} Incorporating self-reflection into inference has been shown to improve model performance \citep{guo2025deepseek}. Self-Refine \citep{madaan2023self} implements an iterative feedback–revision loop without requiring additional training or external supervision. SCoRe \citep{kumar2024training} adopts multi-round reinforcement learning to cultivate reflective reasoning skills, yielding substantial gains. GEPA \citep{agrawal2025gepa} further introduces a Pareto-based reflective prompt-evolution framework that outperforms reinforcement learning in both sample efficiency and overall performance.

Extending self-reflection to multimodal settings likewise yields consistent improvements \citep{yang2025look, jian2025look}. VL-Rethinker \citep{wang2025vl} incorporates a self-verification–and-correction stage via reinforcement learning, substantially enhancing multimodal reasoning. SRPO \citep{wan2025srpo} further strengthens self-reflection and self-correction by curating reflection-oriented supervised data and introducing a reflection-aware reward under group relative policy optimization, significantly boosting complex multimodal reasoning.

However, existing multimodal self-reflection mechanisms still depend on text-based re-evaluation processes. They lack explicit reflective operations over visual information and therefore remain fundamentally text-centric. The REVISOR framework proposed in this paper addresses this limitation by extending text-centric reflection into a truly multimodal reflective process.

\begin{figure}[h]
  \centering
   \includegraphics[width=0.95\linewidth]{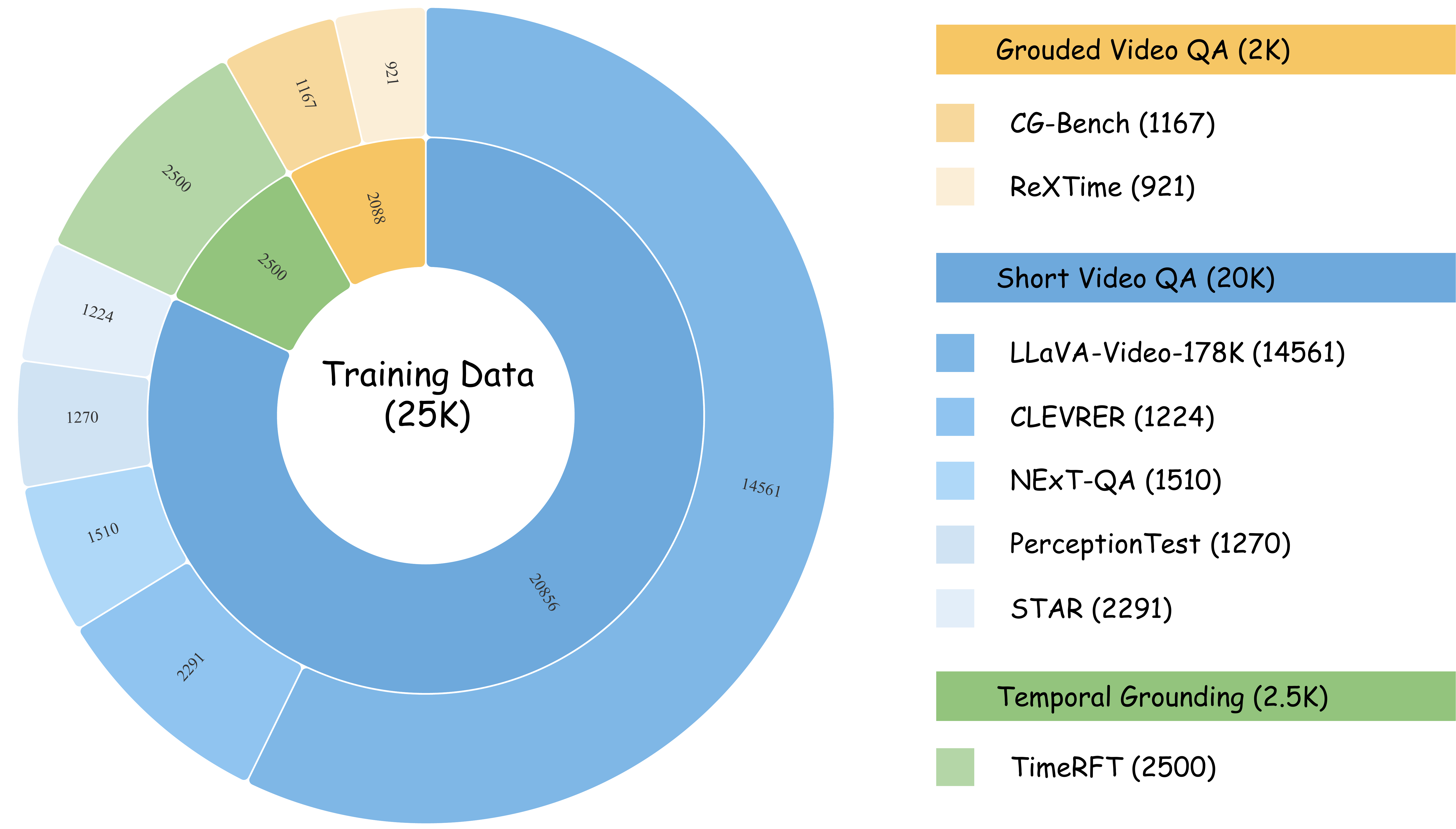}
   \caption{REVISOR framework training dataset composition. The training dataset for the REVISOR framework consists of three tasks: Short Video QA, Temporal Grounding, and Grounded Video QA, totaling 25K training samples. The parenthetical value for each dataset denotes its specific sample contribution.}
   \label{fig: training data}
\end{figure}

\section{More Experimental Details on REVISOR}

In \cref{subapp: experiment setup}, we present the detailed experimental setup for experiments involving the REVISOR framework. \cref{subapp: dataset} outlines the composition of the training data for the REVISOR framework. \cref{subapp: additional ablation} then presents supplementary experimental results, including comprehensive results on the Temporal Video Grounding task and complete ablation studies on the reward scaling factors $\lambda_1$ and $\lambda_2$.

\begin{table*}[t]
\small
\centering
\caption{Comprehensive evaluation of REVISOR on the temporal video grounding task. \textbf{Bold} text indicates the best performance.}
\label{tab:temporal_grounding_all}
\setlength{\tabcolsep}{13pt} 
\begin{tabular}{@{}l|cccc|cccc@{}}
\toprule[1pt]
\toprule
                                 & \multicolumn{4}{c|}{\textbf{Charades-STA}} & \multicolumn{4}{c}{\textbf{NExT-GQA}} \\ 
\cmidrule(lr){2-5} \cmidrule(lr){6-9}
\textbf{Model} & \textbf{R@0.3} & \textbf{R@0.5} & \textbf{R@0.7} & \textbf{mIoU} & \textbf{R@0.3} & \textbf{R@0.5} & \textbf{R@0.7} & \textbf{mIoU} \\ 
\midrule
Qwen2.5-VL-7B\citep{bai2025qwen2} & 57.1 & 33.6 & 15.5 & 36.9 & 31.6 & 18.1 & 7.5 & 20.9 \\
VTimeLLM\citep{huang2024vtimellm} & 55.3 & 34.3 & 14.7 & 34.6 & 37.9 & 20.2 & 9.7 & 24.4 \\
iMOVE\citep{li2025imove} & 71.7 & 51.3 & 26.1 & 47.3 & - & - & - & - \\
TimeChat\citep{ren2024timechat} & 51.5 & 32.2 & 13.4 & - & 34.1 & 17.9 & 6.2 & 20.6 \\
VideoChat-TPO\citep{yan2025task} & 58.3 & 40.2 & 18.4 & 38.2 & 41.2 & 23.4 & 8.2 & 27.7 \\
TVG-R1\citep{chen2025datasets} & 70.8 & 50.5 & 23.9 & 46.7 & 41.7 & 20.8 & 10.0 & 29.3 \\
\rowcolor[HTML]{D8F4FF} 
Ours & \textbf{76.5} & \textbf{57.3} & \textbf{31.8} & \textbf{51.4} & \textbf{47.6} & \textbf{25.5} & \textbf{11.9} & \textbf{33.2} \\ 
\bottomrule
\bottomrule[1pt]
\end{tabular}
\end{table*}

\begin{table*}[t]
\small
\centering
\caption{Complete ablation study of the Dual Attribution Decoupled Reward mechanism. \textbf{Bold} fonts highlight the best performance. The row marked in gray represents our base model.}
\vspace{-0.2cm}
\label{tab:reward_ablation_all}
\setlength{\tabcolsep}{13pt} 
\begin{tabular}{@{}cc|ccccc@{}}
\toprule[1pt]
\toprule
$\boldsymbol{\lambda_1(R_{final})}$ & $\boldsymbol{\lambda_2(R_{causal})}$ & \textbf{VideoMME} & \textbf{LongVideoBench} & \textbf{LVBench} & \textbf{MLVU} & \textbf{NExT-GQA} \\ 
\midrule
\rowcolor[HTML]{EFEFEF} - & - & 64.3 & 56.5 & 40.2 & 67.3 & 20.87 \\ 
\midrule
0.3 & 0.6 & 64.0 & 56.0 & 41.1 & 68.7 & 33.9 \\
0.6 & 0.0 & 62.2 & 54.0 & 40.8 & 68.3 & 32.1 \\
0.45 & 0.45 & 64.3 & 57.3 & 41.2 & 68.5 & 33.7 \\
0.5 & 0.4 & 64.9 & 57.1 & 41.8 & 69.0 & 33.4  \\
\rowcolor[HTML]{D8F4FF} 0.6 & 0.3 & \textbf{65.7} & \textbf{57.5} & \textbf{42.0} & \textbf{69.8} & \textbf{33.2} \\ 
0.8 & 0.1 & 64.6 & 57.6 & 40.3 & 68.4 & 31.6 \\
\bottomrule
\bottomrule[1pt]
\end{tabular}
\end{table*}

\subsection{Detailed Experimental Setup}
\label{subapp: experiment setup}

Our experiments utilized Qwen2.5-VL-7B as the base model, which comprises a visual encoder, a merger projector, and a large language model. REVISOR was trained using a single stage of reinforcement learning, eliminating the need for an additional supervised fine-tuning phase. We extended the verl framework to support REVISOR's training. Following the approach in DAPO, we removed the KL regularization term from GRPO.

Key hyperparameters were set as follows: $\lambda_{1} = 0.6$, $\lambda_{2} = 0.3$, a learning rate of $1 \times 10^{-6}$, a batch size of 32, and 8 rollouts. During both training and evaluation, input video tokens were limited to a maximum of 8192, sampled at 1 FPS. For the Review Segment, the sampling rate was increased to 2 FPS, while still adhering to a maximum of 8192 video tokens. Absolute timestamps were displayed in the lower-left corner of each image frame. The entire training phase consisted of 792 optimization steps. 

\subsection{Detailed Information of Training Data}
\label{subapp: dataset}

As detailed in \cref{fig: training data}, our training corpus spans three tasks central to video understanding: Short Video QA, Temporal Grounding, and Grounded Video QA. For Short Video QA, we aggregate data from LLaVA-Video-178K, CLEVRER, NExT-QA, PerceptionTest, and STAR, all of which provide short video clips ($\leq$ a few minutes). From these datasets, we sample a balanced set of 20K video–question–answer triplets. The Temporal Grounding task is sourced directly from TimeRFT without further filtering. For Grounded Video QA, we incorporate CG-Bench and ReXTime, from which we randomly select 2K examples.

\subsection{Supplementary Experimental Results}
\label{subapp: additional ablation}

\textbf{Temporal Video Grounding.} As summarized in \cref{tab:temporal_grounding_all}, we conduct an extensive assessment of REVISOR’s temporal grounding performance across both the Charades-STA and NExT-GQA benchmarks. On Charades-STA, REVISOR attains an R@0.3 accuracy of 76.5\%, surpassing the prior state-of-the-art method iMOVE by a margin of 4.8\%. On NExT-GQA, REVISOR reaches an R@0.5 accuracy of 25.5\%, outperforming TVG-R1, designed explicitly with reinforcement learning to enhance temporal localization, by 4.3\%. Together, these results highlight the robustness and effectiveness of REVISOR, equipped with DADR, in achieving precise temporal video localization.

\noindent\textbf{Ablation Study of DADR Mechanism.} We conducted an in-depth analysis of the Dual Attribution Decoupled Reward (DADR) mechanism. As shown in \cref{tab:reward_ablation_all}, when $\lambda_2 = 0$, that is, when only the Final Answer Verification Reward is applied, the performance of REVISOR on Video-MME decreases substantially from 65.7\% to 62.2\%, even falling below the base model. This demonstrates that, without the Causal Segment Sufficiency Reward (CSSR), the model struggles to reliably identify the correct review segment $S$ due to the sparsity of the reward signal.

\begin{figure*}[t]
  \centering
   \includegraphics[width=0.95\linewidth]{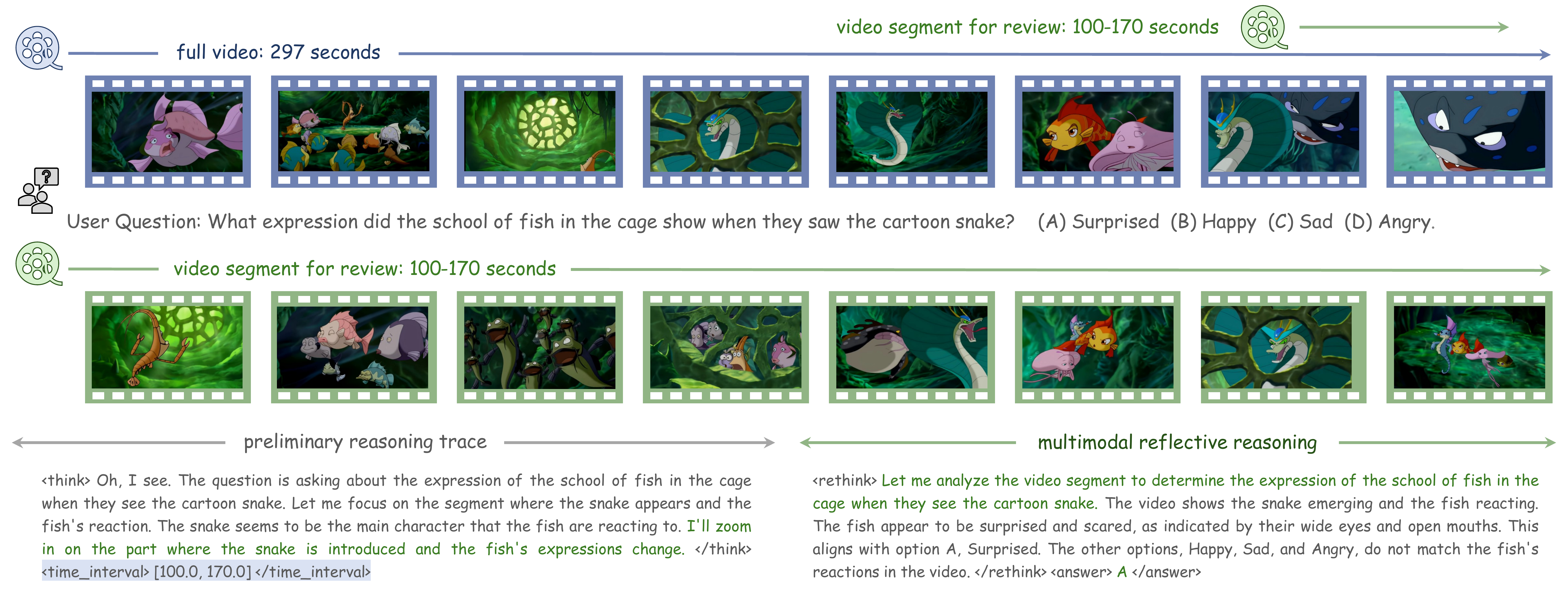}
   \caption{Successful example of the REVISOR framework: achieving more accurate detail capture. By reviewing key video segments, the REVISOR framework accurately identified that the fish in the cage exhibited surprise when they saw the cartoon snake. Without visual reflection, the model mistakenly believed the fish's emotion was anger.}
   \label{fig: case_01}
\end{figure*}

When $\lambda_2$ exceeds $\lambda_1$, the model’s temporal grounding improves; however, its long-video reasoning capability degrades. For example, MLVU performance declines from 69.8\% to 68.7\%, suggesting that the model overemphasizes locating the review segment $S$ while underutilizing it for answer derivation. Consequently, we set $\lambda_1 > \lambda_2$ to encourage accurate localization of $S$ while still promoting strong reasoning based on it.

Nonetheless, if $\lambda_1$ is too large and $\lambda_2$ too small, the model’s ability to locate $S$ deteriorates, harming long-form video understanding. Conversely, when $\lambda_1$ is only slightly greater than $\lambda_2$, REVISOR identifies $S$ effectively but fails to fully leverage it for reasoning, resulting in strong temporal grounding but weaker long-video comprehension. Empirically, we find that $\lambda_1 = 0.6$ and $\lambda_2 = 0.3$ provide an effective balance, enabling REVISOR to both accurately localize $S$ and utilize it to enhance reasoning performance.

\section{Training Qwen2.5-VL-7B with Textual Reflection Mechanism on Video Data}
\label{app: training reflection}

To ensure a fair comparison, we train a Qwen2.5-VL-7B model equipped with a text-based self-reflection mechanism using the datasets listed in \cref{fig: training data}. Specifically, unlike REVISOR, the text-reflection model generates only textual output during the reflection phase. Apart from this distinction, all other experimental settings remain identical to those of REVISOR. The learning rate is set to $1 \times 10^{-6}$, the total batch size is 32, and the number of rollouts is set to 8. In addition, the total number of input video tokens is capped at 8192. The prompt template is shown in \cref{fig:text_rethinking}.

\section{Case Study of REVISOR Framework}
\label{app: case study}

\begin{figure*}[t]
  \centering
   \includegraphics[width=0.98\linewidth]{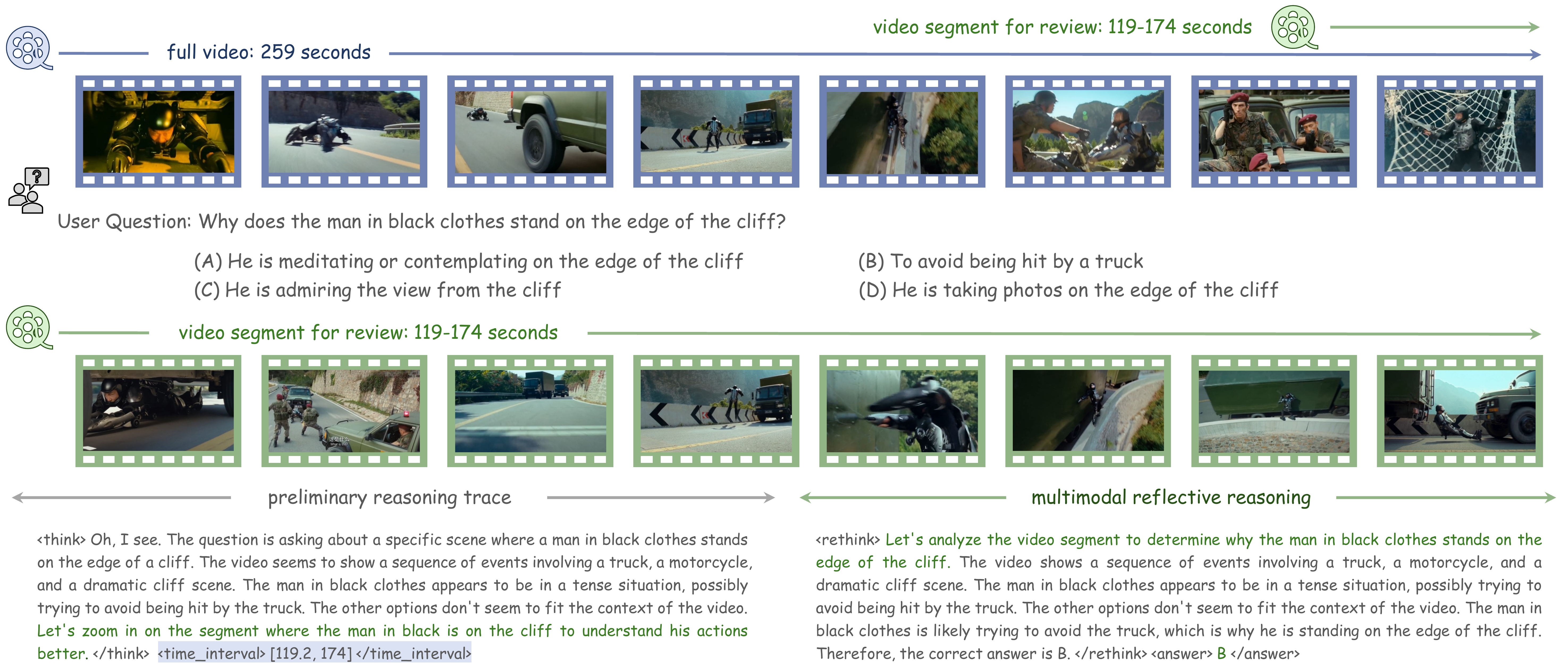}
   \caption{Successful example of the REVISOR framework: achieving more accurate scene understanding. By reviewing key video segments, the REVISOR framework accurately determined that the man was standing on the cliff edge to avoid an oncoming truck. Without this visual reflection, the model erroneously concluded that he was merely taking photos on the cliff.}
   \label{fig: case_02}
\end{figure*}

\begin{figure*}[t]
  \centering
   \includegraphics[width=0.98\linewidth]{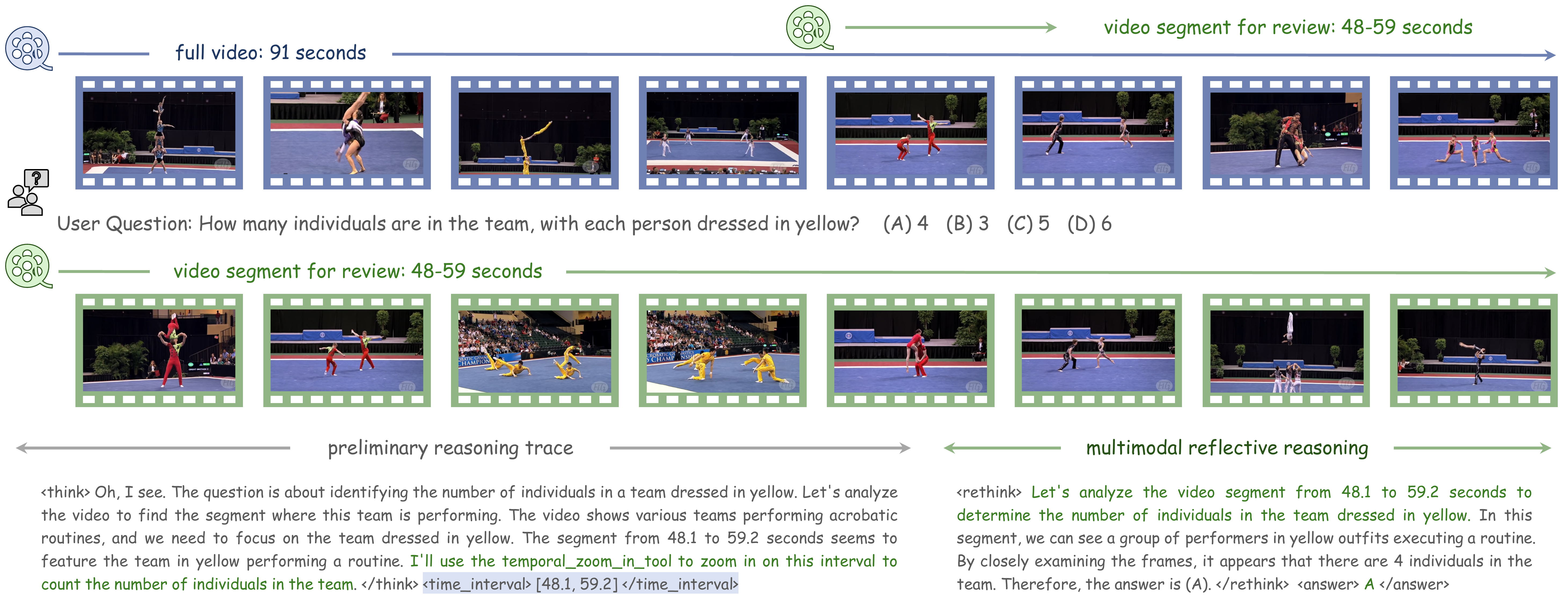}
   \caption{Successful example of the REVISOR framework: achieving more accurate object counting. By reviewing key video segments, the REVISOR framework accurately identified four team members wearing yellow. Without this visual reflection, the model erroneously counted only three.}
   \label{fig: case_03}
\end{figure*}

\textbf{Improved Video Reasoning Capability.} The REVISOR framework can significantly enhance the long-form video understanding capabilities of MLLMs. \cref{fig: case_01}, \cref{fig: case_02}, and \cref{fig: case_03} respectively demonstrate these improvements from three perspectives: more precise detail capture, more comprehensive scene understanding, and more accurate object counting.

\begin{figure*}[t]
  \centering
   \includegraphics[width=0.98\linewidth]{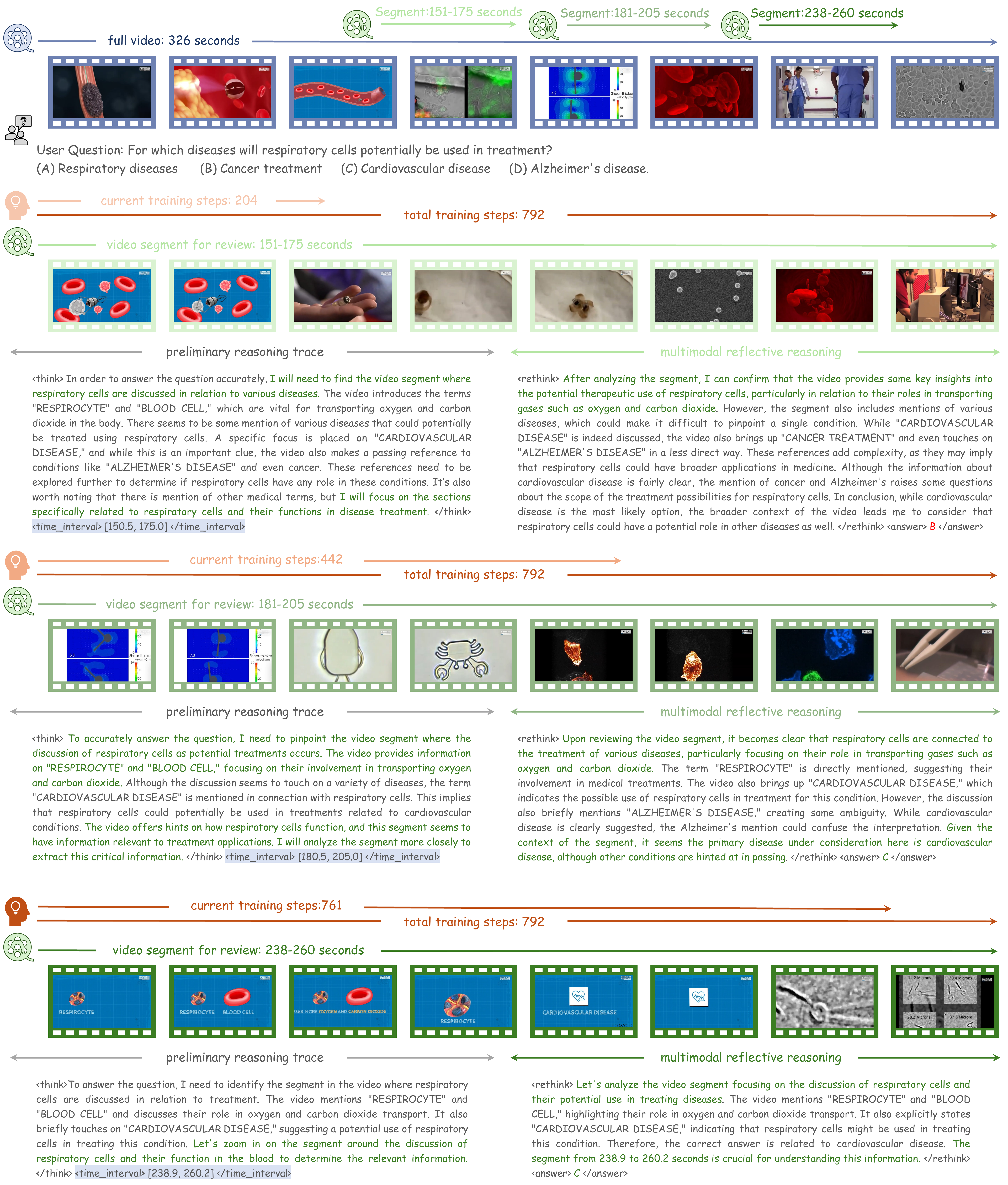}
   \caption{As training progressed, REVISOR's outputted reflective text paragraphs became increasingly concise, and the video segments it referenced grew more precise. Ultimately, it accurately concluded that respiratory cells could potentially be used in the treatment of cardiovascular disease.}
   \label{fig: case_04}
\end{figure*}

\vspace{0.05cm}
\noindent \textbf{Visual Reflection is Even More Important.} In this paper, we repeatedly emphasize that, for long-video understanding tasks, visual reflection is more important than textual reflection. In \cref{subsec:discuss-visual}, we quantitatively validate this conclusion by monitoring changes in the length of textual reflections during training. Here, we further illustrate this point with a concrete example. As shown in \cref{fig: case_04}, during training, the textual reflection generated by the REVISOR framework for a given question becomes increasingly concise, while its retrieval of key video segments becomes increasingly accurate. Ultimately, the model precisely identifies the critical 25-second segment within a 300-second video, confirming that respiratory cells could potentially be used in treatment for cardiovascular disease. This phenomenon is not an isolated incident; similar observations were made across virtually all samples.

\section{Prompt Templates of REVISOR}
\label{app: prompt template}

The complete prompt template used during the training of the REVISOR framework consists of three primary components: the system prompt, the initial reasoning stage, and the reflective reasoning stage. \cref{fig: revisor reasoning templete} illustrates the templates for both the initial reasoning and reflective reasoning stages, while \cref{fig: revisor system templete} presents the system prompt template. A key distinction from the plain-text reflection mechanism is that, during the initial reasoning stage, the REVISOR framework identifies the critical video segments that require further examination.

\begin{figure*}[t]
\centering
\begin{tcolorbox}[
    colback=gray!10,
    colframe=black,
    title={The Complete Prompt Template for Plain-Text Reflection Mechanism},
    width=0.98\textwidth,
    boxrule=0.8pt,
    arc=2mm,
    fonttitle=\bfseries,
    left=4mm, right=4mm, top=2mm, bottom=2mm
]
\textcolor{blue}{\textbf{System Prompt:}}\\[2pt]
You are a helpful assistant.\\[4pt]

\textcolor{red}{\textbf{Stage 1: Preliminary Reasoning Phase}}\\[4pt]
\textcolor{blue}{\textbf{User Instruction:}}\\[2pt]
\texttt{\textless User Question\textgreater }. Please think carefully and then provide your answer.\\[4pt]
\textcolor{blue}{\textbf{Output Format:}}\\[2pt]
Format strictly as: \texttt{\textless think\textgreater} Your reasoning steps \texttt{\textless /think\textgreater} \texttt{\textless answer\textgreater} Your answer \texttt{\textless /answer\textgreater}\\[4pt]

\textcolor{red}{\textbf{Stage 2: Reflective Reasoning Phase}}\\[4pt]
\textcolor{blue}{\textbf{User Instruction:}}\\[2pt]
Please rethink the reasoning process above and provide your final answer.\\[4pt]
\textcolor{blue}{\textbf{Output Format:}}\\[2pt]
Format strictly as: \texttt{\textless rethink\textgreater} Your rethinking reasoning steps \texttt{\textless /rethink\textgreater} \texttt{\textless answer\textgreater} Your final answer \texttt{\textless /answer\textgreater}

\end{tcolorbox}
\caption{The complete prompt template used during the training of Qwen2.5-VL-7B equipped with a pure text-based reflection mechanism, covering both the initial reasoning and reflective reasoning stages.}
\label{fig:text_rethinking}
\end{figure*}

\begin{figure*}[t]
\centering
\begin{tcolorbox}[
    colback=gray!10,
    colframe=black,
    title={Reasoning Prompt Template for REVISOR Framework Training},
    width=0.98\textwidth,
    boxrule=0.8pt,
    arc=2mm,
    fonttitle=\bfseries,
    left=4mm, right=4mm, top=2mm, bottom=2mm
]

\textcolor{red}{\textbf{Stage 1: Preliminary Reasoning Phase}}\\[4pt]
\textcolor{blue}{\textbf{User Instruction:}}\\[2pt]
\texttt{\textless User Question\textgreater }. Please think first, and then use the \texttt{temporal\_zoom\_in\_tool} to find the video segment that can answer the user's question.\\[4pt]
\textcolor{blue}{\textbf{Output Format:}}\\[2pt]
Format strictly as: \texttt{\textless think\textgreater} Your reasoning steps \texttt{\textless /think\textgreater} \texttt{\textless time\_interval\textgreater [start\_time, end\_time] \textless /time\_interval\textgreater}\\[4pt]

\textcolor{red}{\textbf{Stage 2: Reflective Reasoning Phase}}\\[4pt]
\textcolor{blue}{\textbf{User Instruction:}}\\[2pt]
Please refer to the Visual Review segment above, think carefully, and provide your final answer.\\[4pt]
\textcolor{blue}{\textbf{Output Format:}}\\[2pt]
Format strictly as: \texttt{\textless rethink\textgreater} Your reasoning steps \texttt{\textless /rethink\textgreater} \texttt{\textless answer\textgreater} Your final answer \texttt{\textless /answer\textgreater}

\end{tcolorbox}
\caption{Reasoning prompt template for REVISOR framework training}
\label{fig: revisor reasoning templete}
\end{figure*}

\begin{figure*}[t]
\centering
\begin{tcolorbox}[
    colback=gray!10,
    colframe=black,
    title={System Prompt Template for REVISOR Framework Training},
    width=0.98\textwidth,
    boxrule=0.8pt,
    arc=2mm,
    fonttitle=\bfseries,
    left=4mm, right=4mm, top=2mm, bottom=2mm
]

You are a helpful assistant.\\

\textcolor{blue}{\textbf{\# Tools}}\\[4pt]
You may call one or more functions to assist with the user query. You are provided with function signatures within \texttt{<tools></tools>} XML tags:\\

\texttt{<tools>}
\begin{lstlisting}
{
  "type": "function",
  "function": {
    "name": "temporal_zoom_in_tool",
    "description": "Identify the precise time segment in the video that contains enough information to answer the question.",
    "parameters": {
      "properties": {
        "interval": {
          "type": "str",
          "description": "The time range to zoom in on, formatted as 'start_time to end_time'. Timestamps are in seconds."
        }
      },
      "required": ["interval"]
    }
  }
}
\end{lstlisting}
\texttt{</tools>}\\

\textcolor{blue}{\textbf{\# Explanation of "temporal zoom-in"}}\\[4pt]
When you call \texttt{temporal\_zoom\_in\_tool} with \texttt{<time\_interval> [start\_time, end\_time] </time\_interval>}, the tool returns a new sequence of denser video frames sampled from the specified time range (\texttt{[start\_time, end\_time]}) in the original video. This provides higher temporal precision, helping you answer the user's question more accurately.\\

\textcolor{blue}{\textbf{\# How to call a tool}}\\[4pt]
Return the interval directly in XML tags: \texttt{<time\_interval> [12.3, 28.7] </time\_interval>}

\end{tcolorbox}
\captionsetup{justification=centering}
\caption{System prompt template for training the REVISOR framework.}
\label{fig: revisor system templete}
\end{figure*}

\end{document}